\documentclass{article}

\usepackage{arxiv}
\usepackage{color}
\usepackage[utf8]{inputenc} % allow utf-8 input
\usepackage[T1]{fontenc}    % use 8-bit T1 fonts
\usepackage{hyperref}       % hyperlinks
\usepackage{url}            % simple URL typesetting
\usepackage{booktabs}       % professional-quality tables
\usepackage{amsfonts}       % blackboard math symbols
\usepackage{nicefrac}       % compact symbols for 1/2, etc.
\usepackage{microtype}      % microtypography
\usepackage{cleveref}       % smart cross-referencing
\usepackage{lipsum}         % Can be removed after putting your text content
\usepackage{graphicx}
\usepackage{natbib}
\usepackage{doi}
\usepackage{xcolor}

\definecolor{orange}{RGB}{255,204,0}

\title{ChatDiT: A Training-Free Baseline for Task-Agnostic Free-Form Chatting with Diffusion Transformers}

\date{}

\author{
Lianghua Huang\thanks{Corresponding Author}
\And
Wei Wang
\And
Zhi-Fan Wu
\AND
Yupeng Shi
\And
Chen Liang
\And
Tong Shen
\And
Han Zhang
\And
Huanzhang Dou
\And
Yu Liu
\And
Jingren Zhou
\AND
\\
{\large Tongyi Lab}\thanks{Emails: Lianghua Huang, Wei Wang, Zhi-Fan Wu, Tong Shen, Yu Liu, Jingren Zhou \{xuangen.hlh, ww413411, wuzhifan.wzf, st456222, ly103369, jingren.zhou\}@alibaba-inc.com, and Yupeng Shi (shiyupeng.syp@taobao.com). Chen Liang (liangchen2022@ia.ac.cn, Institute of Automation, Chinese Academy of Sciences), Han Zhang (hzhang9617@gmail.com, Shanghai Jiao Tong University) and Huanzhang Dou (hzdou@zju.edu.cn, Zhejiang University) contributed to this work during internships at Tongyi Lab.}
}

\renewcommand{\headeright}{}
\renewcommand{\undertitle}{Technical Report}
\hypersetup{
pdftitle={ChatDiT: A Training-Free Baseline for Task-Agnostic Free-Form Chatting with Diffusion Transformers},
pdfsubject={cs.CV},
pdfauthor={Lianghua~Huang, Wei~Wang, Zhi-Fan~Wu, Yupeng~Shi, Chen~Liang, Huanzhang~Dou, Tong~Shen, Han~Zhang, Yu~Liu, Jingren~Zhou},
pdfkeywords={ChatDiT, Zero-shot task generalization, Diffusion transformers, Image generation},
}

\begin{document}
\maketitle

\begin{abstract}
Recent research \citep{huang2024group, huang2024context} has highlighted the inherent in-context generation capabilities of pretrained diffusion transformers (DiTs), enabling them to seamlessly adapt to diverse visual tasks with minimal or no architectural modifications. These capabilities are unlocked by concatenating self-attention tokens across multiple input and target images, combined with grouped and masked generation pipelines. Building upon this foundation, we present ChatDiT, \textbf{a zero-shot, general-purpose, and interactive visual generation framework} that leverages pretrained diffusion transformers in their original form, requiring no additional tuning, adapters, or modifications. Users can interact with ChatDiT to create interleaved text-image articles, multi-page picture books, edit images, design IP derivatives, or develop character design settings, all through \textbf{free-form natural language across one or more conversational rounds}. At its core, ChatDiT employs a multi-agent system comprising three key components: an \textit{Instruction-Parsing agent} that interprets user-uploaded images and instructions, a \textit{Strategy-Planning agent} that devises single-step or multi-step generation actions, and an \textit{Execution agent} that performs these actions using an in-context toolkit of diffusion transformers. We thoroughly evaluate ChatDiT on IDEA-Bench \citep{liang2024ideabench}, comprising 100 real-world design tasks and 275 cases with diverse instructions and varying numbers of input and target images. Despite its simplicity and training-free approach, ChatDiT surpasses all competitors, including those specifically designed and trained on extensive multi-task datasets. While this work highlights the untapped potential of pretrained text-to-image models for zero-shot task generalization, we note that ChatDiT's Top-1 performance on IDEA-Bench achieves a score of \textbf{23.19} out of 100, reflecting challenges in fully exploiting DiTs for general-purpose generation. We further identify key limitations of pretrained DiTs in zero-shot adapting to tasks. We release all code, agents, results, and intermediate outputs to facilitate further research at \url{https://github.com/ali-vilab/ChatDiT}.
\end{abstract}

% keywords can be removed

\section{Introduction}
\label{sec:introduction}

\begin{figure}
\centering
\includegraphics[width=\textwidth]{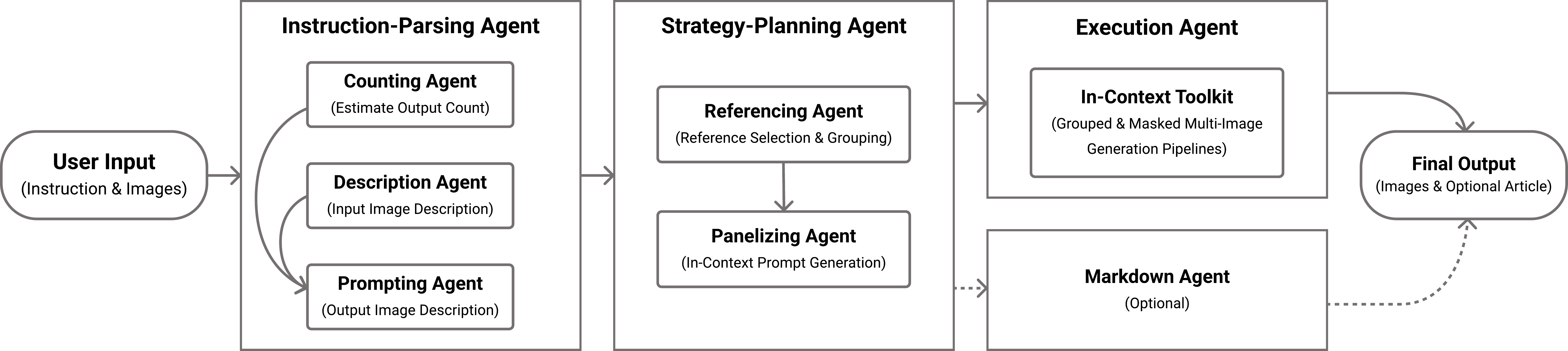}
\caption{\textbf{Overview of the ChatDiT multi-agent framework.} The framework consists of three core agents operating sequentially: the \textit{Instruction-Parsing Agent} interprets user instructions and analyzes inputs, the \textit{Strategy-Planning Agent} formulates in-context generation strategies, and the \textit{Execution Agent} performs the planned actions using pretrained diffusion transformers. An optional \textit{Markdown Agent} integrates the outputs into cohesive, illustrated articles. Sub-agents handle specialized tasks within each core agent, ensuring flexibility and precision in generation.}
\label{fig:fig1}
\end{figure}

Recent advances in text-to-image models have enabled the generation of high-quality images with remarkable fidelity to prompts \citep{ramesh2021zero,esser2021taming,ramesh2022hierarchical,rombach2022high,saharia2022photorealistic,betker2023improving,podell2023sdxl,esser2024scaling,baldridge2024imagen,blackforestlabs_flux_2024}. Additionally, a variety of adapters have been developed to enhance the controllability of these models \citep{zhang2023adding,ye2023ip,huang2023composer,ruiz2023dreambooth,wang2024instantid,hertz2024style}. However, real-world applications often involve complex requirements that surpass the limitations of existing adapters. For instance, generating a picture book necessitates maintaining compositional consistency and intricate variations across a multitude of elements. While recent efforts have sought to develop unified models capable of handling diverse tasks \citep{ge2023making,zhou2024transfusionpredicttokendiffuse,sheynin2024emu,sun2024generative,wang2024emu3}, these approaches typically rely on large amounts of task-specific data and extensive multi-task training. Although such models exhibit zero-shot generalization capabilities, they tend to lack stability on unseen tasks, are challenging to scale, and fail to leverage abundant task-agnostic data effectively.

Emerging work, such as Group Diffusion Transformers \citep{huang2024group}, has proposed a task-agnostic approach by utilizing group data for training. This method allows for the incorporation of diverse relational data sources, such as illustrated articles, video frames, and picture books, making the training data highly redundant. These models demonstrate the potential for zero-shot generalization across various tasks. Building on this, In-context LoRA \citep{huang2024context} simplifies the concept by highlighting the inherent in-context generation capabilities of text-to-image diffusion transformers. By fine-tuning these transformers with a small dataset of 10–100 image groups per task, In-context LoRA achieves impressive results across a range of tasks. However, its reliance on per-task training limits its ability to generalize to unseen tasks.

In this work, we aim to maximize the potential of the core observation underlying In-context LoRA \citep{huang2024context}: that diffusion transformers inherently possess in-context generation capabilities, and consequently, \textbf{zero-shot task generalization potential}. We propose a training-free, zero-shot, interactive, and general-purpose image generation framework built directly upon diffusion transformers in their original form, without the need for fine-tuning, adapters, or structural modifications.

We begin by introducing an \textit{in-context toolkit} for diffusion transformers, enabling them to generate sets of images (instead of single outputs) conditioned on prompts and, optionally, a reference set of images. The toolkit uses a straightforward pipeline similar to that in In-context LoRA, where input and target images are concatenated into a multi-panel layout, described by a comprehensive prompt. The task then involves inpainting the target regions using visible input regions in a training-free manner with blend diffusion \citep{avrahami2022blended}. This pipeline accepts prompts, zero to multiple reference images, and outputs one or more generated images.

The core of our approach, ChatDiT, is a multi-agent system comprising three main agents:

\begin{enumerate}
    \item \textbf{Instruction-Parsing Agent}. This agent interprets user instructions and uploaded images to determine the number of desired output images and generate detailed descriptions for each input and target image.

    \item \textbf{Strategy-Planning Agent}. Based on parsed instructions, this agent formulates a step-by-step generation plan. Each step includes a multi-panel prompt, selected reference image IDs (if applicable), and necessary parameters for image generation.

    \item \textbf{Execution Agent}. Utilizing the \textit{in-context toolkit}, this agent executes the planned steps, generating all target images through in-context operations.
\end{enumerate}

\begin{figure}
    \centering
    \vspace{-12pt}
    \includegraphics[width=\textwidth]{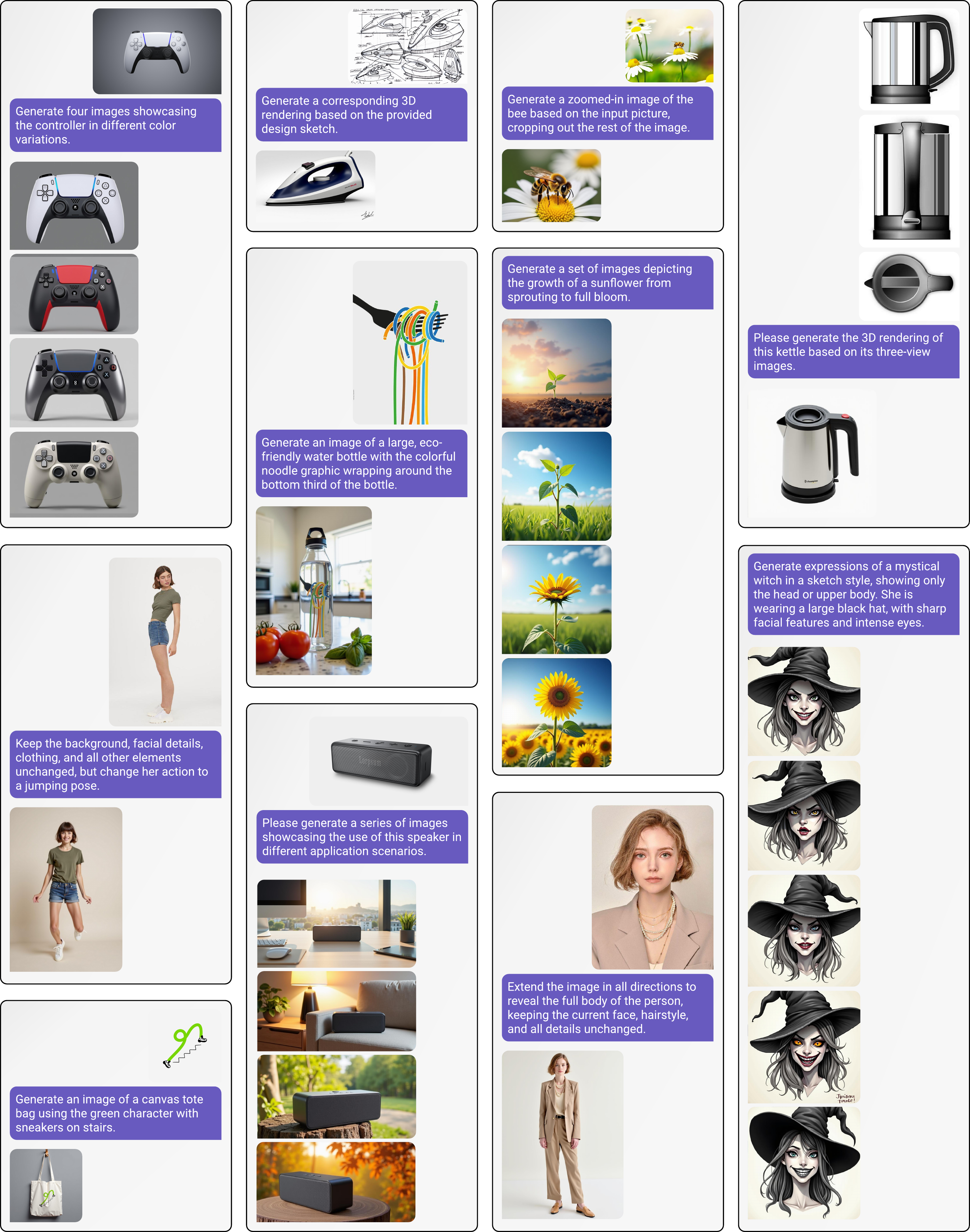}
    \caption{\textbf{Selected single-round generation examples of ChatDiT on IDEA-Bench \citep{liang2024ideabench}.} ChatDiT demonstrates versatility by handling diverse tasks, instructions, and input-output configurations in a zero-shot manner through free-form natural language interactions. The user messages shown here are condensed summaries of the original detailed instructions from \textbf{IDEA-Bench} to conserve space.}
    \label{fig:fig2}
\end{figure}

\begin{figure}
    \centering
    \vspace{-12pt}

    \makebox[\textwidth]{
        \hspace{-30pt}
        \includegraphics[width=1.05\textwidth]{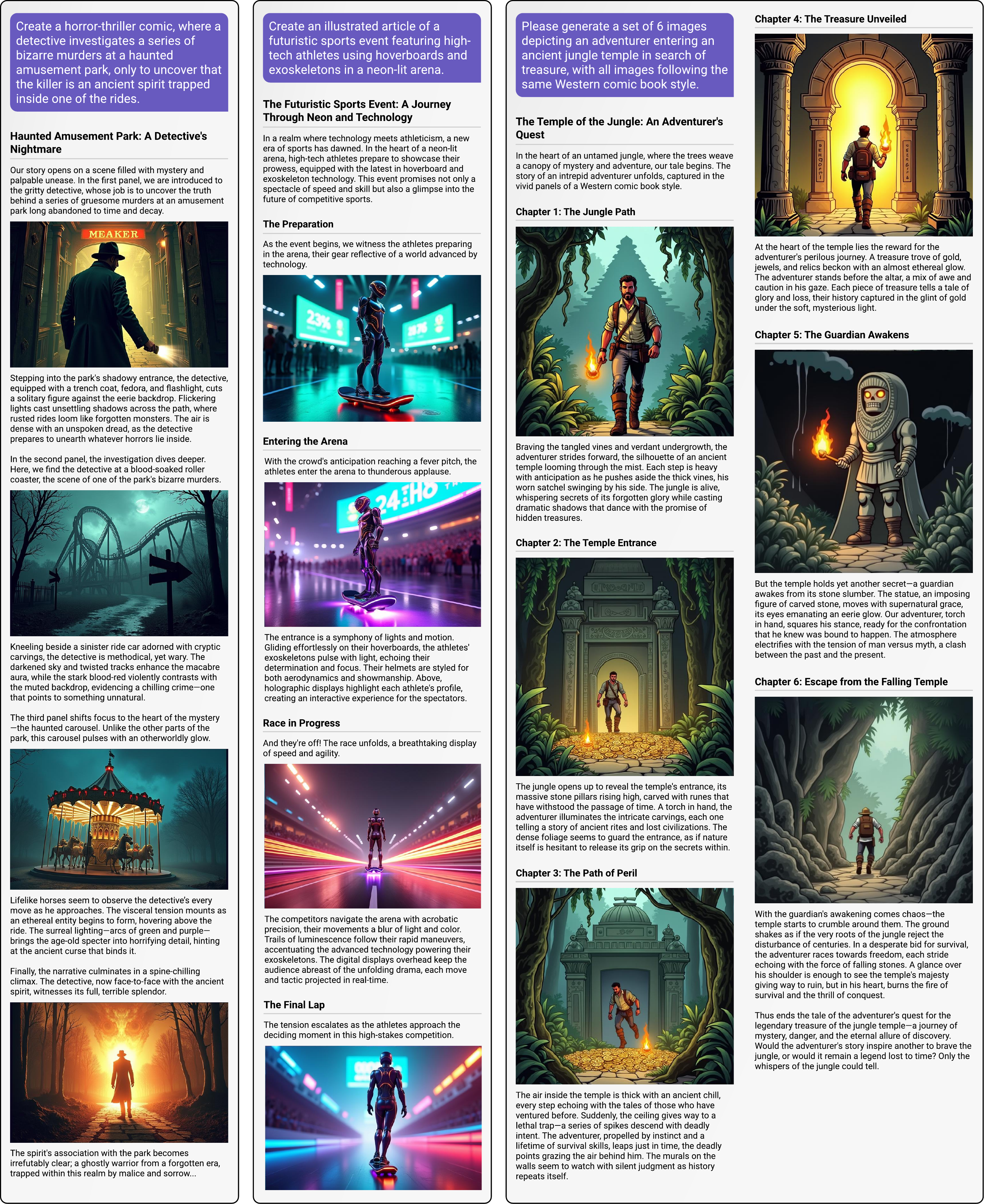}
        \hspace{-30pt}
    }
    
    \caption{\textbf{Selected illustrated article generation examples of ChatDiT.} ChatDiT is able to generate interleaved text-image articles based on users’ natural language instructions. It autonomously estimates the required number of images, plans and executes the generation process using in-context capabilities, and seamlessly integrates the outputs into coherent and visually engaging illustrated articles.}
    \label{fig:fig3}
\end{figure}

\begin{figure}
    \centering
    \vspace{-10pt}

    \makebox[\textwidth]{
        \hspace{-30pt}
        \includegraphics[width=1.05\textwidth]{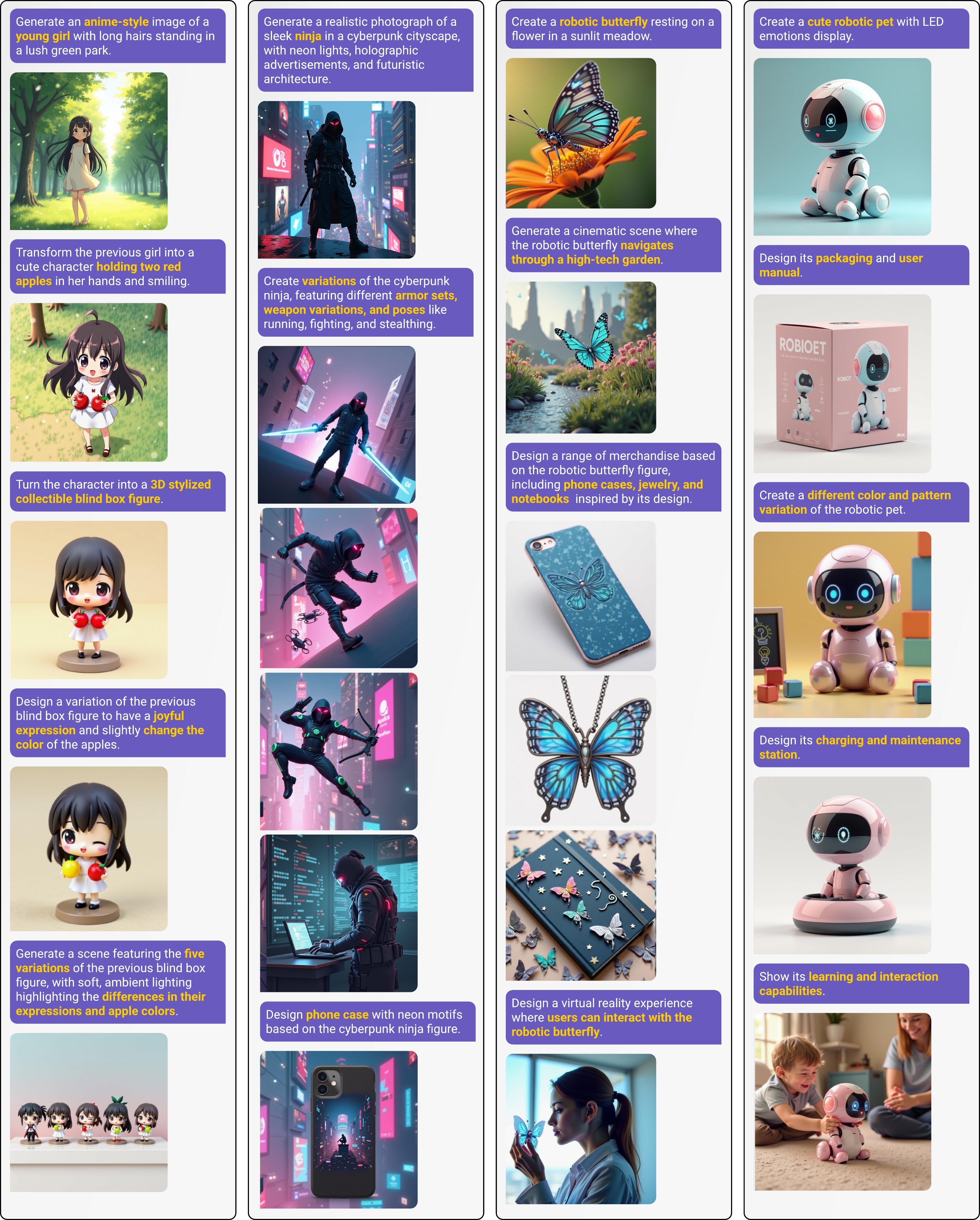}
        \hspace{-30pt}
    }
    
    \caption{\textbf{Selected multi-round conversation examples of ChatDiT.} By referencing images from the conversation history, ChatDiT is able to perform seamless multi-round generation and editing in response to free-form user instructions. This iterative process enables dynamic refinement and adaptation of outputs while maintaining contextual consistency across conversation turns. Key modifications specified in each instructional message are highlighted in \textbf{\textcolor{orange}{yellow}}.}
    \label{fig:fig4}
\end{figure}

An optional \textbf{Markdown Agent} enables the generation of coherent, interleaved text-image articles, ensuring outputs are seamlessly formatted for readability. All agents are implemented using large language models (LLMs) and operate through JSON-based inputs and outputs (except for Markdown Agent outputs, which are text-based). The overall framework is illustrated in Figure \ref{fig:fig1}.

We evaluate ChatDiT on the IDEA-Bench \citep{liang2024ideabench}, a comprehensive benchmark consisting of 100 diverse design tasks and 275 test cases, covering a wide range of instructions and input-output configurations. Example generation results are presented in Figure \ref{fig:fig2}, with quantitative and qualitative comparisons against other approaches shown in Table \ref{tab:tab1} and Figure \ref{fig:fig5}, respectively. Despite its simplicity and training-free nature, ChatDiT outperforms all competitors, including rephrasing-based text-to-image methods and specialized multi-task frameworks, demonstrating its zero-shot capabilities.

We further highlight ChatDiT's versatility in Figure \ref{fig:fig3}, which illustrates its ability to generate interleaved text-image articles, and in Figure \ref{fig:fig4}, which showcases its multi-turn conversational outputs. While some imperfections remain--such as difficulties in identity and detail preservation, and a decline in instruction adherence when handling \textbf{long contexts} (\textit{i.e.,} an excessive number of inputs and/or outputs)--ChatDiT establishes a strong baseline. It also reveals the untapped in-context generation potential of pretrained diffusion models, offering valuable insights into how these models could be further enhanced for improved zero-shot generalization.

Although ChatDiT achieves the best performance on IDEA-Bench, its score of \textbf{23.19} (out of 100) highlights the considerable gap that remains to achieve real-world, product-level general-purpose applications. This result underscores the challenges in fully exploiting the capabilities of diffusion transformers for highly complex tasks. We discuss the key limitations of ChatDiT in Section \ref{sec:limitations}. To foster future research and innovation, we publicly release all code, agents, results, and intermediate outputs\footnote{Project page: \url{https://ali-vilab.github.io/ChatDiT-Page/}}.

\section{Related Work}
\label{sec:related_work}

\subsection{Image Generation}

Text-to-image generation models have rapidly advanced in producing high-fidelity and stylistically diverse images from natural language prompts \citep{ramesh2021zero,ramesh2022hierarchical,esser2021taming,rombach2022high,saharia2022photorealistic,betker2023improving,podell2023sdxl,chen2023pixart,esser2024scaling,baldridge2024imagen,blackforestlabs_flux_2024}. Researchers have introduced various approaches to control specific attributes including identity preservation \citep{huang2023composer,ye2023ip,li2023photomaker,wang2024instantid}, color adaptation \citep{huang2023composer}, style adaptation \citep{hertz2024style,huang2023composer}, spatial composition \citep{zheng2023layoutdiffusion,huang2023composer}, pose guidance \citep{zhang2023adding}, local editing \citep{meng2021sdedit,lugmayr2022repaint,xie2022smartbrushtextshapeguided,huang2023composer}, object-level editing \citep{pan2023drag,shi2023dragdiffusion,liu2024drag}, quality enhancement \citep{saharia2022image,kawar2022denoising,xia2023diffirefficientdiffusionmodel,li2023diffusion}, and cross-image relationship modeling \citep{zhou2024storydiffusion,liu2024intelligent,yang2024seed}. While these methods address individual tasks, they rely on specialized training or adapters, limiting their applicability to broader, more complex tasks that involve multiple images and intricate relationships.

\subsection{Unified Frameworks and Zero-Shot Generalization}

Several recent frameworks strive for generalization across a wide range of generation tasks \citep{ge2023making,zhou2024transfusionpredicttokendiffuse,sheynin2024emu,sun2024generative,wang2024emu3,huang2024group,shi2024seededit}. Emu Edit \citep{sheynin2024emu}, Emu2 \citep{sun2024generative}, Emu3 \citep{wang2024emu3}, TransFusion \citep{zhou2024transfusionpredicttokendiffuse}, Show-o \citep{xie2024showo}, OmniGen \citep{xiao2024omnigen}, and other models demonstrate impressive versatility. Emu3, for example, extends text-to-image to video generation, while OmniGen aims at multi-modal tasks using large-scale training on curated datasets.

Despite their breadth, these models typically rely on either explicit multi-task training or large-scale integration of diverse datasets. In contrast, recent studies \citep{huang2024context} indicate that standard text-to-image diffusion transformers already encode powerful in-context capabilities. In-context LoRA \citep{huang2024context}, for instance, trains small LoRA adapters using a modest number of image groups, revealing the model’s latent ability to handle multiple tasks without large-scale retraining. Our work takes a step further by showing that even without such adaptation, pretrained diffusion transformers can exhibit remarkable zero-shot generalization.

\subsection{Multi-Agent Systems and Interactive Frameworks}

The rise of large language models (LLMs) \citep{radford2019language,brown2020language,touvron2023llama,touvron2023llama2,dubey2024llama,geminiteam2024geminifamilyhighlycapable} has inspired multi-agent architectures that leverage reasoning and planning for complex tasks \citep{durante2024agent,wang2024survey}. Agents can analyze inputs, plan strategies, and execute actions with tools or APIs. While multi-agent reasoning is commonly explored in the language domain, we integrate it into visual generation, using LLM-based agents to parse, plan, and execute multi-step workflows with diffusion transformers. This synergy between reasoning agents and latent diffusion models enables a flexible, conversation-driven interface for complex image generation tasks.

\section{Method}
\label{sec:method}

\begin{table}
\caption{Comparison of ChatDiT with other models across various tasks on IDEA-Bench \citep{liang2024ideabench}. Performance metrics are reported for different task types: T2I (Text-to-Image), I2I (Image-to-Image), Is2I (Image set to Image), T2Is (Text-to-Image set), and Is2Is (Image set to Image set). ``+GPT4o'' indicates that user instructions and uploaded images are reformulated into per-output-image prompts, enabling text-to-image models to generate results. The top two scores for each task are highlighted in \textbf{\textcolor{red}{red}} (best) and \textbf{\textcolor{blue}{blue}} (second best).}
\centering
\tiny
\begin{tabular}{l|ccccccccc|c}
\toprule
\textbf{Task Type} & \textbf{FLUX+GPT4o} & \textbf{DALL-E3+GPT4o} & \textbf{SD3+GPT4o} & \textbf{Pixart+GPT4o} & \textbf{InstructPix2Pix} & \textbf{MagicBrush} & \textbf{Anole} & \textbf{Emu2} & \textbf{OmniGen} & \textbf{ChatDiT} \\
\midrule
T2I  & \textbf{\textcolor{blue}{46.06}} & 24.34 & 24.04 & 14.44 & 0 & 0 & 0 & 17.98 & 21.41 & \textbf{\textcolor{red}{50.91}} \\
I2I  & 12.13 & 6.95  & 10.79 & 7.75  & 17.58 & \textbf{\textcolor{blue}{19.07}} & 0.64 & 7.05  & 8.17  & \textbf{\textcolor{red}{21.58}} \\
Is2I & 4.89  & \textbf{\textcolor{blue}{5.27}}  & 4.69  & 3.48  & 0 & 0 & 0 & \textbf{\textcolor{red}{8.98}}  & 2.77  & \textbf{2.36}  \\
T2Is & 20.15 & 14.36 & \textbf{\textcolor{blue}{21.59}} & 17.46 & 0 & 0 & 1.74 & 0 & 0 & \textbf{\textcolor{red}{27.77}} \\
Is2Is & \textbf{\textcolor{red}{29.17}} & 14.44 & 13.06 & \textbf{\textcolor{blue}{21.39}} & 0 & 0 & 0 & 0 & 0 & \textbf{13.33} \\
\midrule
\textbf{Avg.} & \textbf{\textcolor{blue}{22.48}} & 13.07 & 14.83 & 12.90 & 3.52 & 3.81 & 0.48 & 6.80 & 6.47 & \textbf{\textcolor{red}{23.19}} \\
\bottomrule
\end{tabular}
\label{tab:tab1}
\end{table}

\subsection{Problem Formulation}

\subsubsection{Unified Group Generation Paradigm}

We adopt the image generation paradigm introduced in Group Diffusion Transformers \citep{huang2024group} and In-Context LoRA \citep{huang2024context}, where image generation tasks are formulated as producing a set of $n \geq 1$ target images, conditioned on another set of $m \geq 0$ reference images, alongside a comprehensive prompt describing the combined $(n + m)$ images. This unified formulation is highly versatile, accommodating a wide range of design tasks, including picture book generation, storyboard creation, font design and transfer, identity-preserved generation, pose control, image editing, and IP derivation \citep{huang2024group}.

In this framework, the relationships between reference images and target images are implicitly captured through group-wise consolidated prompts. By concatenating reference and target images into a single multi-panel layout and pairing it with corresponding multi-panel prompts, we can seamlessly perform both reference-based and reference-free tasks. The flexibility of this approach lies in its ability to adapt to diverse task requirements simply by varying the number of panels and the configuration of input and output images.

% This unified generation framework leverages the power of pretrained diffusion transformers for a diverse set of tasks, all within a single structure and in a zero-shot manner.

\begin{figure}
    \centering
    \includegraphics[width=\textwidth]{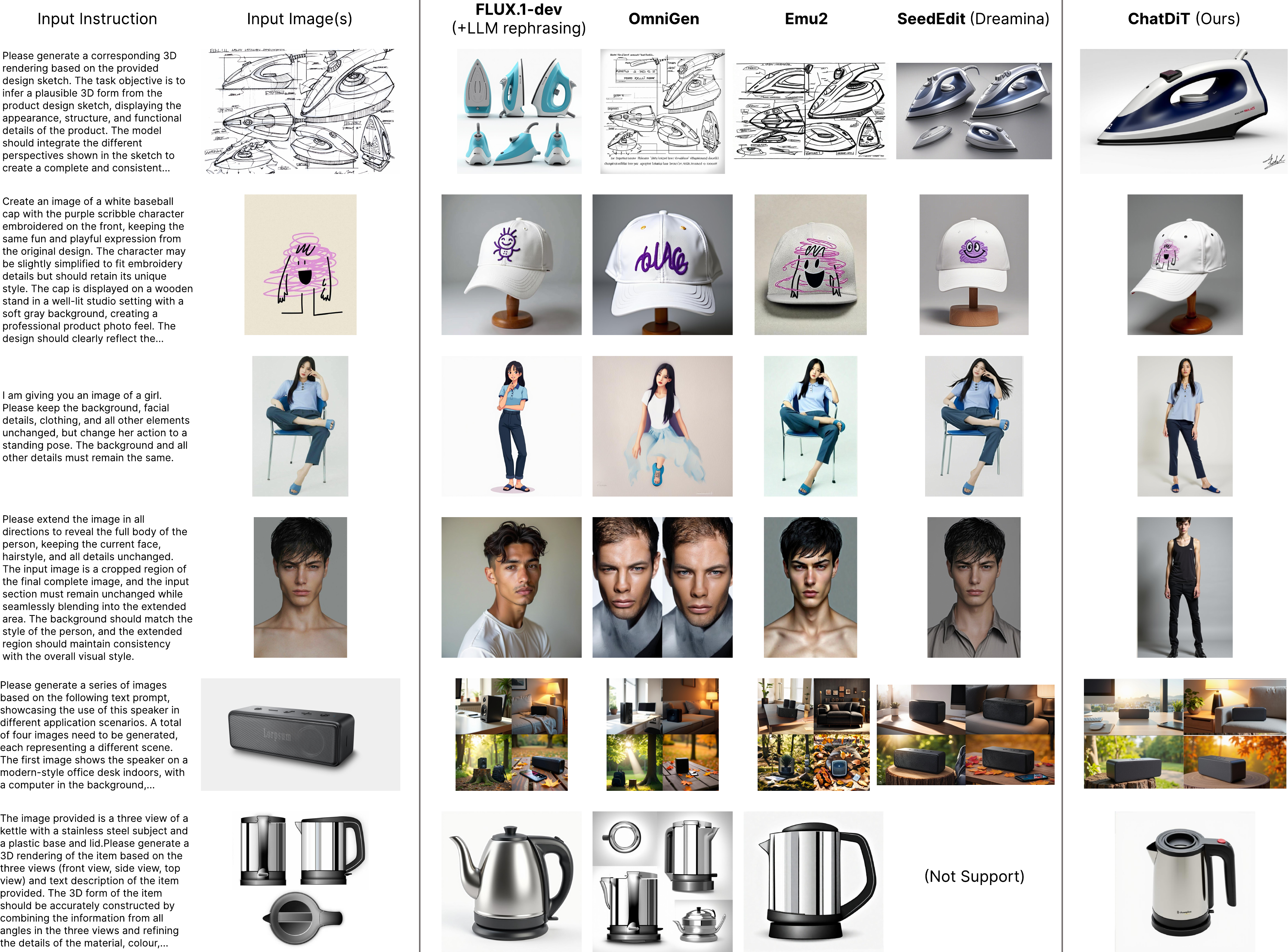}
    \caption{\textbf{Comparison of ChatDiT with existing approaches.}}
    \label{fig:fig5}
\end{figure}

\subsubsection{Alignment with Human Intention}

While the group generation paradigm effectively unifies a broad set of tasks, using multi-panel prompts and image concatenation as the primary interface can be cumbersome. Communicating design requirements through free-form natural language is far more intuitive, \textbf{much like how a consumer would convey ideas to an artist}. Furthermore, when the context is long (i.e., involving many input and/or output images), conditioning or generating all these images simultaneously can significantly degrade performance, as text-to-image models struggle to map multiple panel descriptions accurately.

To address this, we employ a strategy combining parallel and iterative generation actions, preserving relationships between input and target images while maintaining fidelity to image-wise descriptions. The system translates user intent into a format compatible with the in-context toolkit and plans generation strategies that handle large image sets effectively. This involves translating free-form natural language instructions and reference images into structured parameters, devising step-by-step strategies to ensure proper relationships between input and target images, and executing these steps to generate high-quality outputs using the in-context toolkit.

This multi-agent system enables a seamless, user-driven image generation framework that processes natural language instructions and outputs high-quality images, providing a unified, training-free solution to diverse visual generation tasks.

\subsection{In-Context Toolkit}

As demonstrated in previous works \citep{huang2024group,huang2024context}, both reference-free and reference-based multi-image generation tasks can be reformulated as multi-panel image generation and inpainting tasks, which can be effectively handled by pure text-to-image models. In the case of inpainting, a training-free approach is employed \citep{avrahami2022blended}, where the visible regions of the target image are replaced by the corresponding reference image content, with different levels of Gaussian noise added at each denoising step. To ensure accurate image generation, the associated prompt must be comprehensive enough to describe the entire multi-panel content.

To streamline these tasks, we developed an in-context toolkit that integrates essential functionalities such as panel merging and splitting, as well as prompt handling. This toolkit uses a unified interface, simplifying user interaction and enabling seamless integration with the system. Specifically, the toolkit accepts a multi-panel prompt and an image list as inputs and outputs a corresponding image list, expressed as:

\begin{verbatim}
output_images = pipe(prompt, input_images, num_outputs)
\end{verbatim}

This interface is designed for seamless compatibility with the Execution Agent, detailed in the following section.

\subsection{Multi-Agent System}

As illustrated in Figure \ref{fig:fig1}, we have designed a 	\textbf{multi-agent system} to interpret user intentions and generate outputs in a free-form, task-agnostic manner. The system accepts natural language instructions, optionally accompanied by zero or more uploaded images, and produces one or more generated images. When required, the output can be formatted as an illustrated article.

The system comprises three primary agents, each containing specialized sub-agents to handle specific responsibilities:

\begin{itemize}
\item \textbf{Instruction-Parsing Agent}: This agent interprets user instructions and processes the input images. It consists of three sub-agents:
\begin{itemize}
\item \textbf{Counting Agent}: Estimates the number of desired output images based on user instructions.
\item \textbf{Description Agent}: Generates detailed descriptions for each uploaded input image to capture key attributes and context.
\item \textbf{Prompting Agent}: Creates descriptions for the target images to guide the generation process.
\end{itemize}
\item \textbf{Strategy-Planning Agent}: Based on the output from the Instruction-Parsing Agent, this agent formulates a step-by-step generation strategy. It includes:
\begin{itemize}
    \item \textbf{Referencing Agent}: Choose appropriate reference images for each output and organize references and outputs into groups.
    \item \textbf{Panelizing Agent}: Constructs in-context prompts for grouped references and outputs, preparing the input for the image generation pipeline.
\end{itemize}
\item \textbf{Execution Agent}: This agent utilizes the in-context toolkit to execute the generation plan created by the Strategy-Planning Agent, producing the final output images.
\end{itemize}

Additionally, a \textbf{Markdown Agent} is optionally employed to format the generated images and accompanying descriptions into illustrated articles, such as storybooks or instructional content.

Due to the limitations in \textbf{long-context handling} by DiTs, the Strategy-Planning Agent employs specific strategies to optimize the generation process:

\begin{itemize}
\item For \textbf{text-to-images} tasks, the number of panels is capped at 4 to ensure prompt adherence accuracy. If more than 4 outputs are required, subsequent images are generated iteratively, conditioned on the first 3 images.
\item For \textbf{images-to-images} tasks, each output image is generated separately, referencing all input images to ensure consistency.
\item For \textbf{image-to-images} tasks, generation is performed iteratively, conditioning each output on all input images and previously generated outputs.
\end{itemize}

These strategies balance prompt adherence with the need to capture cross-panel relationships and maintain consistency across outputs.

The multi-agent system leverages large language models (LLMs) for the Instruction-Parsing, Strategy-Planning, and Markdown Agents. The Execution Agent employs the in-context toolkit to handle the image generation tasks. JSON-based inputs and outputs are strictly enforced for LLM agents to ensure stability and consistency, except for the Markdown Agent, which outputs markdown-formatted text.

\section{Experiments}

\subsection{Implementation Details}

We utilize the \texttt{FLUX.1-dev} text-to-image model \citep{blackforestlabs_flux_2024} to build the in-context toolkit and the Execution Agent. The large language model (LLM) agents--Instruction-Parsing, Strategy-Planning, and Markdown agents--are implemented using OpenAI's GPT-4o. For the inpainting task, we employ a training-free approach \citep{avrahami2022blended}, using the \texttt{FluxInpaintPipeline} to directly implement panel-wise inpainting for reference-based tasks, ensuring high-quality and contextually accurate image generation.

\subsection{Evaluation Benchmark}

We evaluate the ChatDiT framework using the IDEA-Bench benchmark \citep{liang2024ideabench}, which comprises 100 real-world design tasks, each with varied instructions and different input-output configurations. Spanning 275 cases, the benchmark covers a diverse range of tasks, including picture book creation, photo retouching, image editing, visual effect transfer, and pose transfer.

ChatDiT's performance is compared against several general-purpose frameworks, including OmniGen \citep{xiao2024omnigen}, Emu2 \citep{sun2024generative}, Anole \citep{chern2024anole}, InstructPix2Pix \citep{brooks2023instructpix2pix} and MagicBrush \citep{zhang2024magicbrush}, along with text-to-image models \citep{blackforestlabs_flux_2024,esser2024scaling,chen2023pixart} that incorporate language model rephrasing. These rephrasing-based models translate user-uploaded images and instructions into individual prompts for text-to-image generation. Although such models often fail to capture cross-image relationships, they serve as valuable baselines for comparison, as suggested by IDEA-Bench \citep{liang2024ideabench}.

\subsection{Results on IDEA-Bench}

Table \ref{tab:tab1} presents the quantitative results, while Figure \ref{fig:fig2} provides example generation outputs, and Figure \ref{fig:fig5} visualizes comparisons between ChatDiT and other approaches across selected cases. In terms of overall performance, ChatDiT surpasses competing models, including those explicitly designed and trained on multi-task datasets \citep{xiao2024omnigen,sun2024generative,chern2024anole,brooks2023instructpix2pix,zhang2024magicbrush,shi2024seededit}.

ChatDiT demonstrates strong performance in image-to-image and text-to-image tasks, showcasing its ability to generate high-quality outputs with strong contextual fidelity. However, challenges remain in tasks involving images-to-image and images-to-images scenarios, where extended context length and the complexity of managing multiple inputs and outputs--often with numerous elements or subjects--impact consistency and overall performance.

While ChatDiT exhibits significant capabilities, it still struggles with perfect identity and detail preservation, particularly in human portraits, animal representations, and fine product details. These limitations highlight areas for future enhancement, particularly in maintaining fine-grained visual consistency and accuracy.

\subsection{Interleaved Text-Image Article Generation}

ChatDiT is able to generate interleaved text-image articles by interpreting user instructions alongside input and output image descriptions, converting them into markdown format using the Markdown agent. This process seamlessly integrates text and visuals to produce cohesive, engaging articles. Figure \ref{fig:fig3} highlights a selection of curated examples.

While the current implementation exhibits some imperfections, it demonstrates significant potential for creating interactive and dynamic interfaces. The ability to seamlessly blend text and images paves the way for further enhancements, such as more sophisticated formatting, improved narrative coherence, and expanded functionality in future iterations.

\subsection{Multi-Round Conversation}

Figure \ref{fig:fig4} showcases examples of multi-round conversations using ChatDiT, where the system performs iterative generation and editing based on dynamic, free-form user instructions. By referencing previously generated images and maintaining contextual awareness across conversation turns, ChatDiT is able to refine outputs while preserving consistency and fidelity to user intent.

Although ChatDiT demonstrates promising performance in many cases, challenges remain in preserving fine-grained details and maintaining consistent identity, particularly as the complexity of the conversation increases. Furthermore, cumulative errors can significantly affect performance as the conversation lengthens. Addressing these limitations represents an exciting opportunity for future enhancements.

\subsection{Limitations of ChatDiT}
\label{sec:limitations}

While ChatDiT demonstrates zero-shot generalization capabilities across a variety of visual generation tasks, several limitations remain that highlight areas for further improvement. We summarize these limitations as follows:

\begin{enumerate}
    \item \textbf{Insufficient Reference Fidelity.} ChatDiT struggles to accurately reference details from input images, particularly for maintaining identity and fine-grained details of characters, animals, products, or scenes. While the model can capture overall composition and themes, discrepancies often arise in style consistency, identity preservation, and other nuanced visual attributes.

    \item \textbf{Limited Long-Context Understanding.} The model's performance deteriorates significantly as the number of input or output images increases. When handling long-context scenarios, such as generating large image sets or processing many reference images, ChatDiT's semantic understanding and generation quality drop noticeably, leading to reduced coherence and visual fidelity.

    \item \textbf{Deficiencies in Expressing Narrative and Emotion.} ChatDiT exhibits limited capability in generating content with strong narrative flow, emotional depth, or story-like qualities. This shortcoming can be attributed to the inherent challenges of text-to-image models in capturing and expressing emotions or complex story-driven scenes. Additionally, the model tends to simplify complex scenes, favoring the generation of visually straightforward outputs.

    \item \textbf{Weak High-Level In-Context Reasoning.} ChatDiT has difficulty performing advanced in-context tasks. For example, when provided with a group of input-output image pairs and a new input, the model often fails to infer the desired action or generation task. This limitation highlights the model's current inability to generalize higher-order relationships or abstract reasoning across in-context examples.

    \item \textbf{Limited Handling of Multi-Subject or Multi-Element Complexity.} ChatDiT struggles to manage scenarios involving multiple subjects or elements, such as interactions between characters, crowded scenes, or objects with intricate relationships. In such cases, the generated outputs often lose compositional consistency, resulting in incoherent or incomplete representations.
\end{enumerate}

Addressing these limitations will require advancements in fine-grained reference alignment, long-context comprehension, narrative and emotional generation, and improved reasoning capabilities in in-context settings. These findings provide a foundation for future research aimed at enhancing the general-purpose capabilities of diffusion transformers.

\section{Conclusion and Discussion}

In this paper, we presented ChatDiT, a novel zero-shot, general-purpose, and interactive visual generation framework based on pretrained diffusion transformers. Leveraging the inherent in-context generation capabilities of diffusion models, ChatDiT allows users to seamlessly create complex multi-image outputs, edit images, generate interleaved text-image articles, and design character settings--all with minimal user input and no additional fine-tuning or architectural modifications. By incorporating a multi-agent system, we enable a high degree of flexibility and customization, handling user instructions in natural language and transforming them into structured, step-by-step generation plans.

Despite ChatDiT's zero-shot capabilities, several limitations remain. These include challenges with long-context handling, where performance drops as input-output complexity increases, and issues with fine-grained detail preservation, particularly in human faces, animals, and intricate designs. Additionally, ChatDiT struggles with high-level reasoning and generating narratives with emotional depth. Future improvements should focus on enhancing long-context understanding, fine-tuning for specific domains, and improving reasoning across complex scenarios. Addressing these will expand ChatDiT's potential for more nuanced and consistent visual generation across diverse tasks.

\newpage

\bibliographystyle{unsrtnat}
\bibliography{references}

% \appendix
% \section{Appendix A}

\end{document}